\documentclass[10pt, twocolumn, letterpaper]{article}

\usepackage[pagenumbers]{cvpr} %

\newcommand{\TODO}[1]{\textbf{\color{red}[TODO: #1]}}

\usepackage{multirow}
\newcommand{\ours}{\textsc{Spectrify}\xspace}

\newif\ifshownotes
\shownotesfalse
\ifshownotes
  \newcommand{\plan}[1]{{\color{NavyBlue}#1}}
  \newcommand{\ask}[1]{\textbf{\color{Orange}[ASK: #1]}}

\else
  \newcommand{\plan}[1]{}
  \newcommand{\ask}[1]{}

\fi

\usepackage[table]{xcolor}
\usepackage{booktabs,multirow,graphicx}
\usepackage{enumitem}
\usepackage{subcaption}

\definecolor{cvprblue}{rgb}{0.21,0.49,0.74}
\usepackage[pagebackref, breaklinks, colorlinks, allcolors=cvprblue]{hyperref}

\def\confName{CVPR}
\def\confYear{2027}

\title{\ours: Framework for Hyperspectral Image Generation Using Computer Graphics}

\author{
Yasin Mamedov$^{1,2}$ \quad
Daniil Vladimirov$^{1,2,3}$ \quad
Sofiia Dorogova$^{1,2}$ \\
Bakhtiyar Khalfin$^{1,2}$ \quad
Egor Ershov$^{1,2,3}$ \\[4pt]
$^1$MIRAI, Moscow, Russia \quad $^2$Color Reproduction and Synthesis Institute, Moscow, Russia \\
$^3$AXXX, Moscow, Russia
}

\begin{document}
\twocolumn[{%
\renewcommand\twocolumn[1][]{#1}%
\maketitle
\begin{center}
    \includegraphics[width=\textwidth]{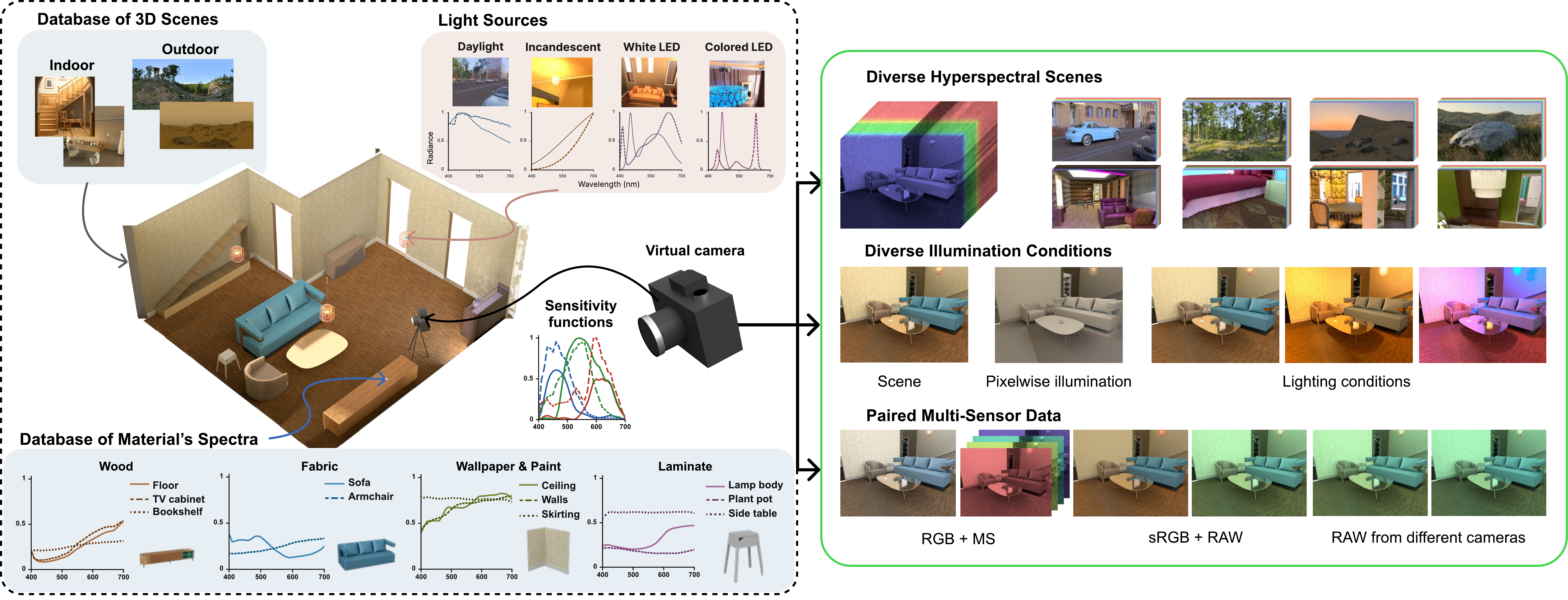}
    \captionof{figure}{\textbf{Overview of \ours.}
        \ours renders existing indoor and outdoor 3D scenes into hyperspectral images.
        Each object part receives measured reflectance spectra of its material category from our library (bottom left), and each lamp and the sky an emission spectrum such as daylight, incandescent or LED light (top).
        Path tracing renders the scene, and projecting the resulting hyperspectral image through the spectral sensitivities of a camera gives that camera's RAW image.
        Spectral rendering thus makes it possible to synthesize training data for many applications, from hyperspectral images and per-pixel illumination maps to aligned images of different sensors (right).}
    \label{fig:abstract}
\end{center}

}]
\begin{abstract}

Hyperspectral images (HSIs) record the full spectrum of light at every pixel, so projecting an HSI through a camera's spectral sensitivities gives that camera's RAW image, and one HSI can supply training data for hyperspectral reconstruction and camera-pipeline methods.
Real HSIs, however, are slow and expensive to capture, and the largest datasets hold only about 1,000 images.
High-level vision commonly replaces scarce real data with images rendered from 3D scenes.
Low-level tasks, however, need physically correct spectra, which requires spectral rendering with a measured spectrum for every material in a scene, and existing spectral renderings cover only a handful of scenes, mostly with a simplified image formation model.
We present \ours, a framework that converts 3D scene datasets into HSIs, together with a library of reflectance spectra labeled by material and compiled from close-range hyperspectral datasets and spectrophotometer measurements.
In \ours, spectra follow the material: each object receives a physically based optical model and measured reflectance spectra or optical constants from its material category, each lamp an emission spectrum of its type and the sky a modeled daylight spectrum, and Mitsuba~3 path-traces the full light transport of each scene per wavelength.
We show that a hyperspectral reconstruction network trained only on our synthetic HSIs generalizes to unseen real datasets better than networks trained on the largest real HSI datasets on most test sets, and that assigning spectra by material category instead of by color alone improves this transfer on every test set.
RAW images projected from our HSIs are also useful training data for color constancy, performing on par with existing synthetic RAW generators.
We release the 4,500 rendered HSIs, the code of \ours and its material-labeled reflectance library.
\end{abstract}

\section{Introduction}
\label{sec:intro}

Learned low-level vision methods require large training sets, and for methods that operate on RAW images these sets are tied to a specific camera.
The RAW values a camera records are determined by its optical system, i.e.\ its lens, filters and sensor, whose combined spectral response is described by the spectral sensitivity functions (SSFs).
Cameras with different optical systems therefore have different SSFs and record different RAW values for the same scene and illumination~\cite{seo2023graphics2raw}, so a model trained on data from one camera degrades on another. %
Training data must therefore be collected for each camera, by capturing the same scenes again with every new device~\cite{kim2026rawgen}.
For phone cameras, which change every year, this collection has to be repeated for each new model.

To avoid recapturing data for every camera, prior work synthesizes RAW training data from existing sRGB images by inverting the image signal processing (ISP) pipeline, an approach known as unprocessing~\cite{brooks2019unprocessing,seo2023graphics2raw}.
Each ISP step is undone with a simple parametric operation, which brings an sRGB image back to the linear RAW space of a target camera.
These inverses assume a single, known pipeline, whereas real sRGB images carry diverse and unknown photo-finishing, so the inversion does not reliably recover the scene-referred linear signal~\cite{kim2026rawgen}.
RawGen~\cite{kim2026rawgen} instead learns the mapping with a diffusion model that turns sRGB images, or images generated from text prompts, into linear CIE XYZ, which calibration matrices then convert to a camera's RAW space.
A diffusion prior, however, can hallucinate content~\cite{kim2024tackling,ren2025hallucination}, and three channels cannot resolve metamerism: spectra that look identical in XYZ can differ for the camera~\cite{jiang2013camspec}.

Hyperspectral images give a more controlled source of sensor data.
An HSI keeps the wavelength-resolved light of a scene, so projecting it through the SSFs of any camera simulates that camera's RAW image~\cite{vora2001,jiang2013camspec}, and one HSI serves every sensor with known SSFs~\cite{kaya2019spectral}.
Real HSIs, however, are even scarcer~\cite{beyond-rgb,harvard-dataset,cave-dataset,arad2016sparse,arad2020ntire,arad2022ntire,mobile-spec,kaust-dataset,reutsky2025modulate,du2025automatic} than RAW data: scanning hyperspectral cameras need static scenes and can take from tens of seconds to minutes per capture~\cite{chakrabarti2011statistics,kim2012imaging}.
ARAD 1K~\cite{arad2022ntire}, one of the largest natural-scene datasets and a standard benchmark for hyperspectral reconstruction, contains about a thousand images, which is small as a training set for modern learned methods~\cite{zhang2022survey,fu2025limitations}.
Because hyperspectral capture is costly, such datasets are usually recorded with a single camera in one region, so they cover a narrow range of content and lighting.

Spectral rendering can remove this bottleneck.
High-level vision already relies on rendered datasets with exact per-pixel labels~\cite{roberts2021hypersim,yang2024depth,ros2016synthia,zhang2017physically}.
A physically based renderer that traces light at every wavelength, such as Mitsuba~3~\cite{framework/mitsuba3}, can likewise produce HSIs of any 3D scene, with spectral shading, interreflections, and mixed lighting~\cite{hao2020mist}.
What such a renderer needs, and what 3D scene datasets lack, is a spectrum for every material, texture, and light source.
Graphics2RAW~\cite{seo2023graphics2raw} mentions spectral rendering as an alternative way to synthesize RAW images but considers it significantly more complicated, since it needs spectra for the whole scene and the SSFs of the target camera.
Rendered synthetic HSI datasets therefore remain small, with spectra assigned by color from a limited set~\cite{hao2020mist} or per material in a few purpose-built scenes~\cite{sippel2023hyvid, lee2026swr}.
Assigning spectra by color, however, is not enough, because surfaces of the same color can have different spectra~\cite{foster2006frequency}, and a color match can then pick the spectrum of the wrong material.

In this work we present \ours, a framework that generates HSIs by physically based spectral rendering of existing 3D scene datasets.
It reads the material information stored in 3D assets and uses it to select measured reflectance spectra from a large material-labeled library that we compile, so no spectrum is assigned by hand and no reflectance is inferred from color alone.
Lamps and skies receive physically plausible emission spectra, and spectral path tracing renders each scene into an HSI, from which aligned RAW images of any camera with known SSFs follow (\cref{fig:abstract}).
Models trained on this data transfer to real images on two downstream tasks, hyperspectral reconstruction and color constancy.

In summary, our contributions are:
\begin{itemize}
\item \ours, a framework that converts existing 3D scene datasets into HSIs by physically based spectral rendering and uses material categories to assign optical models and select measured reflectance spectra.

\item An evaluation of the rendered data on hyperspectral reconstruction and on color constancy from RAW images.
For hyperspectral reconstruction, models trained on our data generalize to unseen real datasets better than models trained on prior synthetic data and, on most test sets, better than models trained on the largest real HSI datasets, and assigning spectra by material category improves this transfer.
For color constancy, RAW images projected from our HSIs perform on par with existing synthetic RAW generators.

\item A release of 4,500 HSIs of indoor and outdoor scenes rendered with \ours, together with the code of \ours and its material-labeled reflectance library.
\end{itemize}

\section{Related Work}
\label{sec:related}

\textbf{Synthetic Data Generators}\quad
Most synthetic training data for low-level vision is derived from existing sRGB images.
UPI~\cite{brooks2019unprocessing} inverts the ISP pipeline to turn sRGB images into RAW, and Graphics2RAW~\cite{seo2023graphics2raw} maps graphics images to the RAW space of a specific camera using its DNG metadata.
Buzzelli \etal~\cite{buzzelli2023pipeline} apply the same idea to spectra and replace each sRGB pixel with the most similar reflectance from a set such as Munsell~\cite{parkkinen1989munsell} or RidiQulous~\cite{ridiqulous}.
These methods assume one known photo-finishing pipeline; RawGen~\cite{kim2026rawgen} removes this assumption by training a diffusion model to map differently photo-finished sRGB versions of a scene to one linear CIE XYZ image.
All of them, however, infer RAW values or spectra from the three channels of an existing image.

Rendering 3D scene datasets~\cite{roberts2021hypersim} avoids this dependence on existing images and provides diverse ground truth~\cite{cogo2026dense}, but the spectra that spectral rendering needs are laborious to prepare for each dataset~\cite{seo2023graphics2raw,hao2020mist}.
Previous spectral generators therefore either choose spectra regardless of material or assign them per material in only a handful of scenes.
MIST~\cite{hao2020mist} gives each color the spectrum closest in chromaticity from a small set of measured spectra.
HyViD~\cite{sippel2023hyvid} crops textures by hand from a hyperspectral dataset~\cite{arad2016sparse}, and in SWR~\cite{lee2026swr} a coding agent builds a few worlds whose spectra come from the material-labeled ECOSTRESS~\cite{ecostress} and USGS~\cite{usgs} libraries.
\ours gives every surface spectra of its own material automatically, so it converts whole 3D scene datasets instead of a handful of scenes.

\textbf{Spectral Rendering}\quad
Most rendering engines aim at visually realistic images and compute light transport in a tristimulus color space, which approximates products of spectra by products of three channel values and so cannot be physically accurate for every combination of light and material~\cite{todova2022wide,todova2025constrained}.
Physically based engines PBRT~\cite{pharr2016physically}, ART~\cite{wilkie2018advanced}, Hydra~\cite{frolov2018hydra}, and Mitsuba~3~\cite{framework/mitsuba3} can instead trace light at sampled wavelengths and output spectral radiance.
These engines solve light transport but leave the spectra of materials and light sources to the user.
We use Mitsuba 3 because all its material models support spectral rendering and its Python API lets us build scenes from code.

\textbf{Reflectance Libraries}\quad
Measured reflectance spectra have been collected for color research, the study of metamerism, and remote sensing.
Collections from color research range from the 1,995 surface spectra of Barnard \etal~\cite{barnard2002}, including Munsell chips~\cite{parkkinen1989munsell}, to the 25M distinct reflectances that Zhang \etal~\cite{zhang2016metamer} gather to study metamer mismatching and the 43M spectra of RidiQulous~\cite{ridiqulous}, with reduced versions of 7M and 114K.
These collections do not label their spectra by material, so a spectrum can be drawn from them only by its color.
USGS~\cite{usgs} and ECOSTRESS~\cite{ecostress} label their spectra by material and SOCS~\cite{tajima2002socs} by object type, but these libraries are small and cover few material classes.
Existing libraries are thus either large or labeled by material, but not both.
Our library is both large and labeled by material.

\section{The \ours Framework}
\label{sec:framework}

\ours takes a 3D scene, with the arrangement and textures of its objects and the placement of its light sources, and gives every object a spectral material and every light source an emission spectrum.
The material category of each object part follows from the object it belongs to.
This category selects both the optical model of the part and the pool of measured spectra it can receive, and the texture then selects spectra within that pool.

\begin{figure*}[t]
  \centering
  \includegraphics[width=\textwidth]{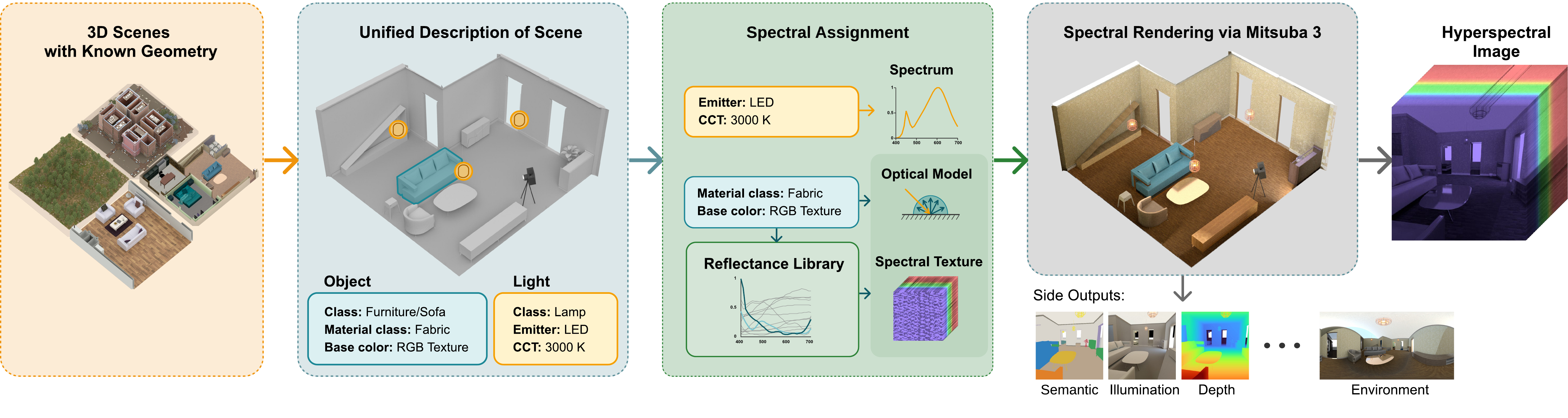}
  \caption{\textbf{The \ours pipeline.}
  A 3D scene is read into a description of its objects and light sources.
  Each object receives a material category, which fixes its optical model and the pool of library spectra it can receive, and its texture color selects spectra within that pool.
  Each light source receives an emission spectrum, and Mitsuba~3 renders a 31-band hyperspectral image by spectral path tracing.}
  \label{fig:pipeline}
\end{figure*}

\Cref{fig:pipeline} shows the pipeline.
We first compile a library of measured reflectance spectra labeled by material category (\cref{sec:library}).
We then determine material categories from the object and material names of a 3D scene and assign optical models, and every light source of the scene receives an emission spectrum (\cref{sec:invscene}).
Each surface then receives reflectance spectra from its category that match its source colors (\cref{sec:textures}).
Finally, Mitsuba~3 renders the scene by spectral path tracing (\cref{sec:rendering}).

\subsection{A Material-Labeled Reflectance Library}
\label{sec:library}

Each material category needs enough labeled spectra to cover the colors it takes in 3D scenes.
Close-range hyperspectral images provide such numbers but are mostly unlabeled, while spectrophotometer measurements are labeled but few, so we use both: about 16,600 close-range hyperspectral images and 29K spectrophotometer measurements from 70 datasets.
We take 63 material categories from existing material-labeled datasets~\cite{framework/opensurfaces,framework/matsynth,framework/ambientcg,framework/polyhaven}.
Spectrophotometer measurements and annotated hyperspectral datasets come with labels that we map to these categories.
Unannotated hyperspectral images are rendered to RGB and segmented with Segment Anything Model~3~\cite{framework/sam3}, using the category names as text prompts.
Each labeled pixel of a hyperspectral image and each measurement gives one spectrum, about 370M in total.
All spectra are resampled to the 31 rendering bands (\cref{sec:rendering}).
Spectra are stored as absolute reflectance in $[0,1]$, not normalized to unit peak, so a material keeps its brightness as well as its spectral shape.
Library statistics and the list of source datasets are given in \cref{supp:library}.

\subsection{Material and Light Assignment}
\label{sec:invscene}

This step turns the objects and lamps of a source scene into materials and light sources for spectral rendering: each object part receives a material category, which fixes its optical model, and each lamp receives an emission spectrum.
We apply it to 3D-FRONT~\cite{framework/3dfront} interiors and to city and natural scenes that we build from third-party asset collections~\cite{framework/polyhaven,framework/ambientcg}.

\textbf{Material categories}\quad
Objects can consist of several parts with different materials, such as a chair with a wooden frame and a fabric seat, so each part receives its own category, read from its material name and the object name through a table that maps the names of each dataset to our categories.
When the names give no category, it is sampled from a prior over the materials of the object class; for example, a chair is wood with probability 0.5, fabric with 0.3, and metal with 0.2.

\textbf{Optical models}\quad
The category also fixes the BSDF model, since how a surface scatters light depends on what it is made of.
Most categories use a rough coated model, a diffuse base under a dielectric coating, and matte categories (\eg plaster, fabric) use a diffuse model.
In both models, the measured reflectance spectrum sets the diffuse component, and in the coated model the specular reflection follows from the refractive index of the coating alone.
Metals use a rough conductor with wavelength-dependent refractive index and extinction coefficient from measurements of 39 metals~\cite{refractiveindex}, so their color comes from the measured optical constants instead of a reflectance spectrum.
Transparent categories use a rough dielectric for closed meshes and a thin dielectric for open ones, with a scalar refractive index from measurements or set by category.

\textbf{Light sources}\quad
We use the lamps of the source scenes and assign each a lamp type: white and colored LEDs, fluorescent, incandescent, halogen, or high-pressure discharge.
White LEDs span correlated color temperatures from 2700 to 7000\,K.
Colored LED lamps use measured spectra of narrowband LED chips~\cite{framework/brendel-led}.
Lamp power follows from a target illuminance chosen by room type through the lumen method~\cite{framework/lumen-method}.
Daylight comes from a catalogue of spectral sun and sky radiance maps computed with libRadtran~\cite{framework/libradtran} over solar positions and atmospheric conditions, including aerosols and clouds.

\subsection{Spectral Textures from RGB}
\label{sec:textures}

Textures carry much of the visual detail of 3D scenes, such as wood grain, fabric patterns, and print, so we convert each RGB texture into a spectral texture instead of giving the surface a single spectrum.
The conversion has two requirements: every texel's spectrum is made of measured spectra of the surface's material category, and the color differences between texels, such as gradients and patterns, are preserved.
We give each texel a blend of measured spectra of its material in three steps:
\begin{enumerate}[label=(\roman*),nosep,leftmargin=*]
\item compute the CIE XYZ colors of the category's library spectra under D65 and divide them into tetrahedra with a Delaunay triangulation;
\item find the tetrahedron that contains the texel color and write this color as a weighted sum of its four corner colors, with non-negative weights that sum to one;
\item give the texel the same weighted sum of the four corner spectra.
\end{enumerate}
The color of a weighted sum of spectra is the same weighted sum of their colors, so the texel keeps its color, and its spectrum consists only of real spectra of its material.
The blend weights also change continuously with the color, so texels of similar color, such as neighboring texels along wood grain, receive similar spectra, so gradients and patterns are kept.
A texel color that the material never takes in our measurements, such as a saturated blue on wood, lies outside all tetrahedra and is moved to the nearest one, which keeps the material at the cost of some color accuracy.
The appearance of the texture under the light sources of a scene then follows from these spectra, not from the RGB values.

\subsection{Rendering}
\label{sec:rendering}

For a non-emissive surface point $\mathbf{x}$, the spectral radiance toward the camera is
\begin{equation}
L_o(\mathbf{x},\omega_o,\lambda) = \int
f_s(\mathbf{x},\omega_i,\omega_o,\lambda)\,
L_i(\mathbf{x},\omega_i,\lambda)\,
|\cos\theta_i|\,
\mathrm{d}\omega_i,
\label{eq:framework/spectral_rendering}
\end{equation}
where $L_i$ is the incident spectral radiance from direction $\omega_i$, $\omega_o$ is the viewing direction, and $\theta_i$ is the angle between $\omega_i$ and the surface normal.
The BSDF $f_s$ is the model assigned to the material category in \cref{sec:invscene}, with the spectral reflectance or the measured optical constants entering it per wavelength.
Because the integral is evaluated at every wavelength, interreflections between colored surfaces and mixtures of lamps and daylight are spectrally correct.
\begin{figure}[!ht]
  \centering
  \setlength{\tabcolsep}{1pt}
  \begin{tabular}{cc}
    \small Real & \small Rendered \\
    \includegraphics[width=0.49\linewidth]{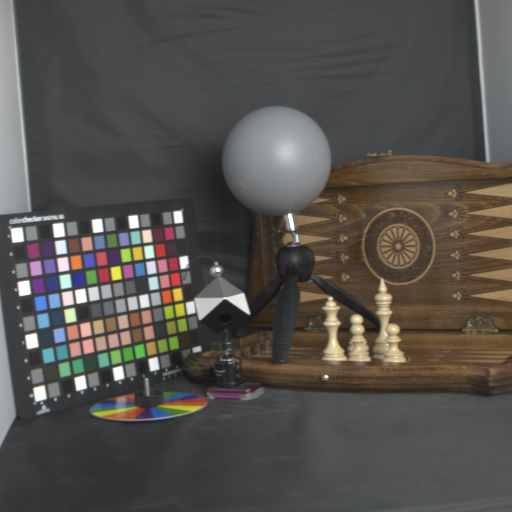} &
    \includegraphics[width=0.49\linewidth]{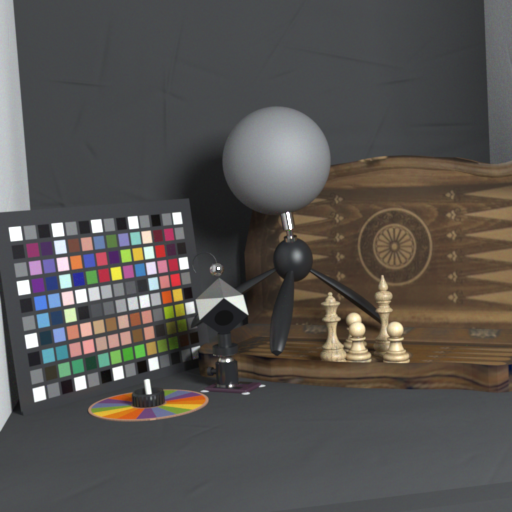}
  \end{tabular}
  \caption{\textbf{A real scene and its rendering.}
  A photograph of a tabletop scene with a color chart, a gray sphere and wooden objects (left) and our rendering of its 3D model, with spectra assigned by hand (right).}
  \label{fig:twin}
\end{figure}

We evaluate it with the spectral path tracer of Mitsuba~3~\cite{framework/mitsuba3} in the 31 bands of our library, from 400 to 700\,nm at 10\,nm steps, at $512\times512$ pixels, the resolution of the widely used Specim~IQ hyperspectral camera~\cite{behmann2018specim}.
Indoor scenes are rendered with 10,240 samples per pixel and outdoor scenes with 1024, which takes about 3 minutes per scene on one NVIDIA A100 GPU with about 15\,GB of GPU memory.

\begin{figure*}[!ht]
    \centering
    \includegraphics[width=\textwidth]{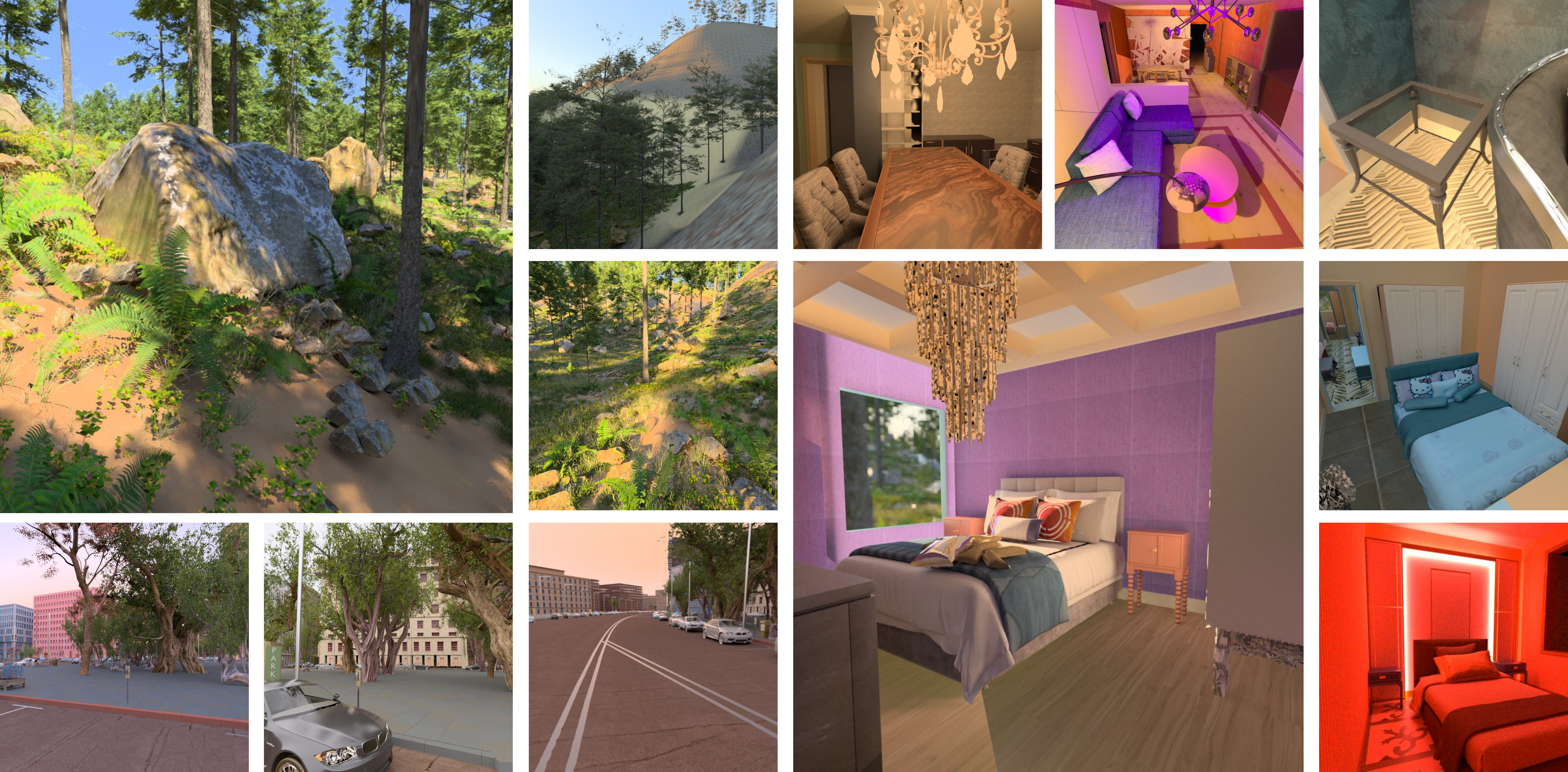}
    \caption{\textbf{Scenes rendered with \ours, shown in sRGB.}
    Natural, city and indoor scenes under different lighting conditions.}
    \label{fig:render_examples}
\end{figure*}

\ours can also output semantic labels, depth, environment maps and illumination maps.
\Cref{fig:twin} shows a photograph of a real tabletop scene next to the rendering of its 3D model, whose spectra we assign by hand, and \cref{fig:render_examples} shows representative scenes rendered with \ours.

\section{Experiments}
\label{sec:experiments}

We evaluate \ours on two downstream tasks that use our rendered HSIs in different ways.
In hyperspectral reconstruction, a network recovers an HSI from a RAW image, so the HSIs serve as targets as well as, after projection through a camera's SSFs, as inputs; this task tests the rendered spectra themselves.
In color constancy, a network estimates the scene illuminant from a RAW image, so only the projection of the spectra is used.
Both tasks use one set of 4,500 rendered HSIs with the same training, validation and test split.
Hyperspectral reconstruction networks trained on one dataset generalize poorly to others, because each dataset covers few scenes~\cite{fu2025limitations}.
We therefore test generalization to four real datasets unseen during training, comparing our data with real data and with the 7,022 synthetic HSIs of Buzzelli \etal~\cite{buzzelli2023pipeline}, and measure the contribution of material-based spectral assignment by replacing it with assignment by color alone.
For color constancy, we test on NUS~\cite{cheng2014nus} and Cube++~\cite{ershov2020cube}, two real datasets that both contain images from the Canon EOS 600D, and compare our data with RawGen~\cite{kim2026rawgen}, which produces RAW images without modeling spectra, and with Buzzelli \etal, whose HSIs are projected to RAW through the same SSFs as ours.

\subsection{Cross-Dataset Generalization of Hyperspectral Reconstruction from RGB}
\label{sec:exp-hsi}

Hyperspectral reconstruction (HSR) networks learn from paired RGB images and HSIs, and existing datasets of such pairs cover few scenes, illuminants and capture devices.
This limited diversity is the main obstacle to applying HSR to real-world images~\cite{arad2020ntire,zhang2022survey,fu2025limitations}.

We use four real datasets that differ in content and lighting.
ARAD~1K~\cite{arad2022ntire} and ICVL~\cite{arad2016sparse} were collected by the same group and contain mostly daylight scenes.
BJTU-UVA~\cite{du2025automatic} contains urban and natural scenes captured under different weather, lighting and times of day.
Doomer~\cite{reutsky2025modulate} mixes indoor scenes lit by halogen and narrow-band LED lamps with outdoor scenes under mostly overcast skies.
Each dataset serves both as a training source and as a test set.
As synthetic training sources we use \ours and the pipeline of Buzzelli \etal~\cite{buzzelli2023pipeline}, which assigns measured reflectance spectra to the pixels of sRGB photographs.
We remove the spectra of all four real datasets from our reflectance library (\cref{sec:library}) before rendering, so the library contains no spectra from the test sets.

All HSIs are projected to RAW RGB through the spectral sensitivities of a Nikon D700 camera and corrupted with the same noise model, and every source trains the same MST++ network~\cite{cai2022mstpp} (details in \cref{supp:hsr}).
A network trained on one source is tested on the test split of every real dataset.
When the training and test sets come from the same dataset, the result is in-domain and serves as the reference for that test set.
We report PSNR, SAM and the normalized spectral error (NSE)~\cite{reutsky2025modulate}.

\providecommand{\indom}{\cellcolor{gray!15}} %
\providecommand{\best}[1]{\textbf{#1}}
\providecommand{\second}[1]{\underline{#1}}
\begin{table*}[!ht]
  \centering
  \caption{\textbf{Cross-dataset generalization of hyperspectral reconstruction.}
    MST++ is trained on each source (rows) and tested on the test split of each real dataset (column groups), with all HSIs projected to RAW through the spectral sensitivities of a Nikon D700.
    Shaded cells are in-domain results, which serve as the reference for each test set and are not ranked.
    Bold marks the best and underline the second-best out-of-domain result per column.
    Training on \ours gives the best out-of-domain result on most test sets and metrics and the second best on the rest.
    Arrows show whether higher or lower is better.}
  \label{tab:cross_dataset}
  \setlength{\tabcolsep}{3pt}
  \renewcommand{\arraystretch}{1.6}
  \resizebox{\textwidth}{!}{%
  \begin{tabular}{c l ccc ccc ccc ccc}
    \toprule
    & \multirow{2}{*}{Train set}
      & \multicolumn{3}{c}{ARAD~1K}
      & \multicolumn{3}{c}{Doomer}
      & \multicolumn{3}{c}{ICVL}
      & \multicolumn{3}{c}{BJTU-UVA} \\
    \cmidrule(lr){3-5} \cmidrule(lr){6-8} \cmidrule(lr){9-11} \cmidrule(lr){12-14}
      & & PSNR, dB$\uparrow$ & SAM, $^\circ$$\downarrow$ & NSE, \%$\downarrow$
        & PSNR, dB$\uparrow$ & SAM, $^\circ$$\downarrow$ & NSE, \%$\downarrow$
        & PSNR, dB$\uparrow$ & SAM, $^\circ$$\downarrow$ & NSE, \%$\downarrow$
        & PSNR, dB$\uparrow$ & SAM, $^\circ$$\downarrow$ & NSE, \%$\downarrow$ \\
    \midrule
    \multirow{4}{*}{\rotatebox[origin=c]{90}{Real}} & ARAD~1K
      & \indom 40.44 & \indom 3.90 & \indom 7.87
      & 27.88 & 13.42 & 25.31
      & \second{37.47} & \best{3.74} & \best{5.57}
      & 30.53 & 10.08 & 15.68 \\
    & Doomer
      & \second{33.61} & \second{8.71} & \second{13.11}
      & \indom 35.90 & \indom 4.54 & \indom 8.21
      & 36.29 & 4.65 & 6.69
      & 31.42 & \second{9.40} & \second{15.01} \\
    & ICVL
      & 32.99 & 11.32 & 23.46
      & 27.14 & 14.03 & 30.15
      & \indom 43.83 & \indom 1.71 & \indom 3.10
      & \best{32.86} & 9.45 & 19.02 \\
    & BJTU-UVA
      & 30.98 & 11.50 & 15.29
      & 26.50 & 13.84 & 22.93
      & 33.33 & 7.67 & 10.13
      & \indom 42.67 & \indom 2.11 & \indom 4.72 \\
    \midrule
    \addlinespace[4pt]
    \multirow{2}{*}[-2pt]{\rotatebox[origin=c]{90}{Synthetic}} & Buzzelli
      & 29.98 & 12.79 & 16.85
      & \second{28.08} & \second{11.29} & \second{18.27}
      & 33.97 & 6.50 & 9.47
      & 27.16 & 16.21 & 25.25 \\[4pt]
    & Ours
      & \best{36.99} & \best{6.54} & \best{10.54}
      & \best{28.19} & \best{11.21} & \best{18.19}
      & \best{37.55} & \second{4.13} & \second{6.12}
      & \second{31.51} & \best{8.79} & \best{13.69} \\
    \addlinespace[4pt]
    \bottomrule
  \end{tabular}%
  }
\end{table*}

\Cref{tab:cross_dataset} shows that real datasets transfer poorly to each other: a network trained on one real dataset loses several decibels of PSNR on any other.
Training on our data gives the best out-of-domain result on most test sets and metrics, and the second best on the rest.

\begin{figure*}[!ht]
  \centering
  \begin{subfigure}{\textwidth}
    \centering
    \includegraphics[width=\textwidth]{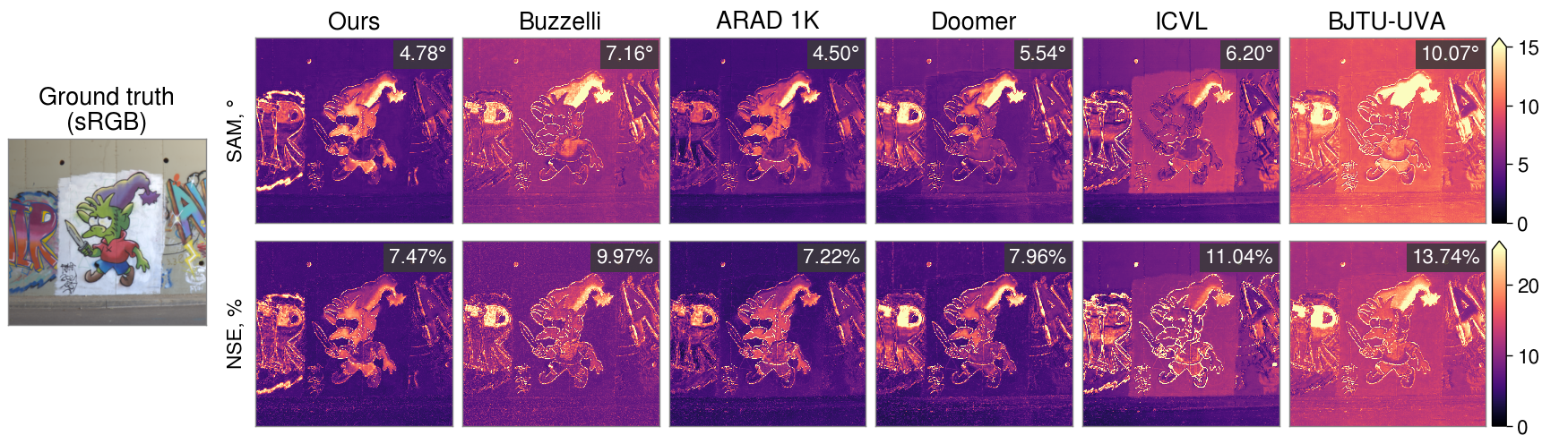}
    \label{fig:error_maps_arad}
  \end{subfigure}

  \begin{subfigure}{\textwidth}
    \centering
    \includegraphics[width=\textwidth]{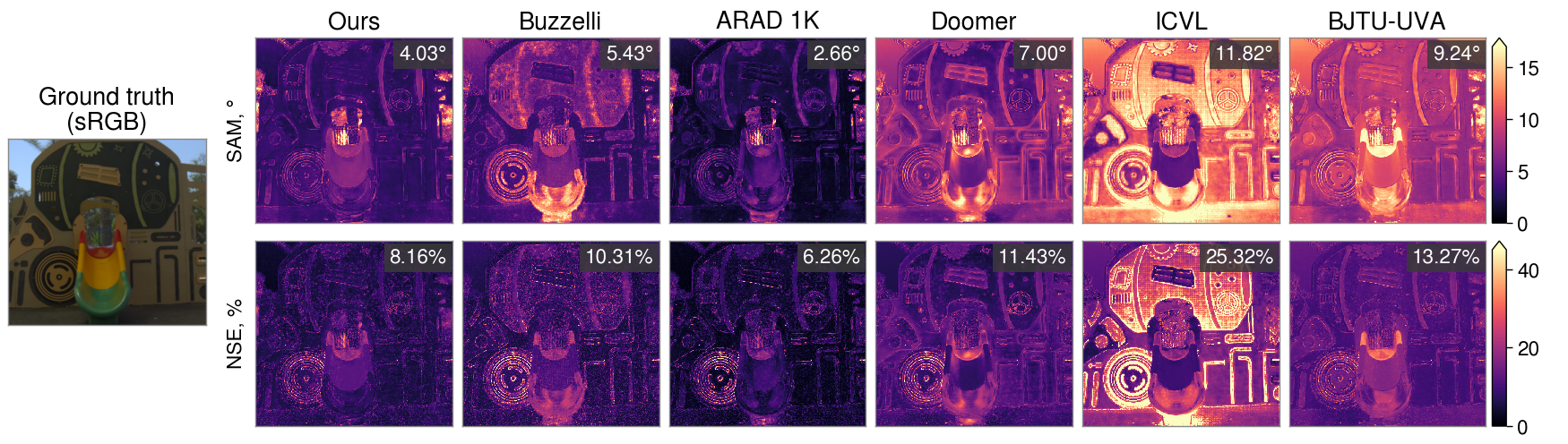}
    \label{fig:error_maps_icvl}
  \end{subfigure}

  \caption{Per-pixel reconstruction error of MST++ trained on each dataset, on two ARAD~1K test images.
    For each image, the ground-truth sRGB render is followed by the spectral angle (SAM, upper row) and the normalized spectral error (NSE, lower row) of each network.
    The ARAD~1K column is the in-domain reference.
    Each row shares one color scale, darker is better, and the corner value is the image mean.
    Among out-of-domain networks, training on our data gives the lowest error on both images.}
  \label{fig:error_maps}
\end{figure*}

On ICVL, training on ARAD~1K gives a lower SAM and NSE than training on our data, while our PSNR is slightly higher.
The two datasets come from the same group and contain similar scenes, so for ICVL, ARAD~1K is close to in-domain data.

Networks trained on real data transfer worst to Doomer, whose LED-lit indoor scenes have little or no counterpart in the other real datasets.
Both synthetic sources transfer to Doomer better than any real source, and both include narrow-band LED spectra, which may contribute to this.
On the other three test sets, however, Buzzelli \etal\ fall behind most real sources, while our data stays first or second.
Since everything except the HSIs is shared, part of this gap may come from how spectra are assigned to surfaces, which we test in \cref{sec:ablation-material}.
Results for other HSR methods are given in the supplementary material (\cref{supp:hsr}).
\Cref{fig:error_maps} shows per-pixel errors on two ARAD~1K test images.

\subsection{Effect of Material-Based Spectral Assignment}
\label{sec:ablation-material}
    To validate that spectra should be assigned by material category (\cref{sec:invscene}), we compare two versions of \ours that differ only in how surfaces receive their spectra.
With the default material + color assignment, the spectrum of each texel is matched to its color among the library spectra of the surface's material category.
With color-only assignment, it is matched over the whole library, as in Buzzelli \etal~\cite{buzzelli2023pipeline}.
We follow the same protocol as in \cref{sec:exp-hsi}.
\Cref{tab:hsr-material-ablation} shows that material + color assignment improves transfer on every test set.
The gain is marginal on ARAD~1K and substantial on the other three.

\providecommand{\best}[1]{\textbf{#1}}
\providecommand{\second}[1]{\underline{#1}}
\begin{table*}[!ht]
  \centering
  \caption{\textbf{Effect of material-based spectral assignment.}
    MST++ is trained on two versions of \ours that differ only in how spectra are assigned to surfaces and evaluated as in \cref{tab:cross_dataset}.
    Color only matches the color of each texel over the whole library, while material + color matches it only among the library spectra of the surface's material category.
    Material + color assignment, the default in \ours, improves transfer on every test set (better result per column in bold).}
  \label{tab:hsr-material-ablation}
  \setlength{\tabcolsep}{3pt}
  \renewcommand{\arraystretch}{1.6}
  \resizebox{\textwidth}{!}{%
  \begin{tabular}{c ccc ccc ccc ccc}
    \toprule
    \multirow{2}{*}{{\renewcommand{\arraystretch}{1}\begin{tabular}[c]{@{}c@{}}Spectral\\Assignment\end{tabular}}}
      & \multicolumn{3}{c}{ARAD~1K}
      & \multicolumn{3}{c}{Doomer}
      & \multicolumn{3}{c}{ICVL}
      & \multicolumn{3}{c}{BJTU-UVA} \\
    \cmidrule(lr){2-4} \cmidrule(lr){5-7} \cmidrule(lr){8-10} \cmidrule(lr){11-13}
      & PSNR, dB$\uparrow$ & SAM, $^\circ$$\downarrow$ & NSE, \%$\downarrow$
      & PSNR, dB$\uparrow$ & SAM, $^\circ$$\downarrow$ & NSE, \%$\downarrow$
      & PSNR, dB$\uparrow$ & SAM, $^\circ$$\downarrow$ & NSE, \%$\downarrow$
      & PSNR, dB$\uparrow$ & SAM, $^\circ$$\downarrow$ & NSE, \%$\downarrow$ \\
    \midrule
    Color only
      & 36.38 & 6.66 & 10.88
      & 26.36 & 13.57 & 21.92
      & 31.92 & 7.60 & 10.86
      & 29.15 & 11.08 & 15.77 \\
    \shortstack{Material + color\\(default)}
      & \best{36.99} & \best{6.54} & \best{10.54}
      & \best{28.19} & \best{11.21} & \best{18.19}
      & \best{37.55} & \best{4.13} & \best{6.12}
      & \best{31.51} & \best{8.79} & \best{13.69} \\
    \bottomrule
  \end{tabular}%
  }
\end{table*}

\subsection{Color Constancy}
\label{sec:exp_awb}

\begin{figure*}[!ht]
    \centering
    \includegraphics[width=\textwidth]{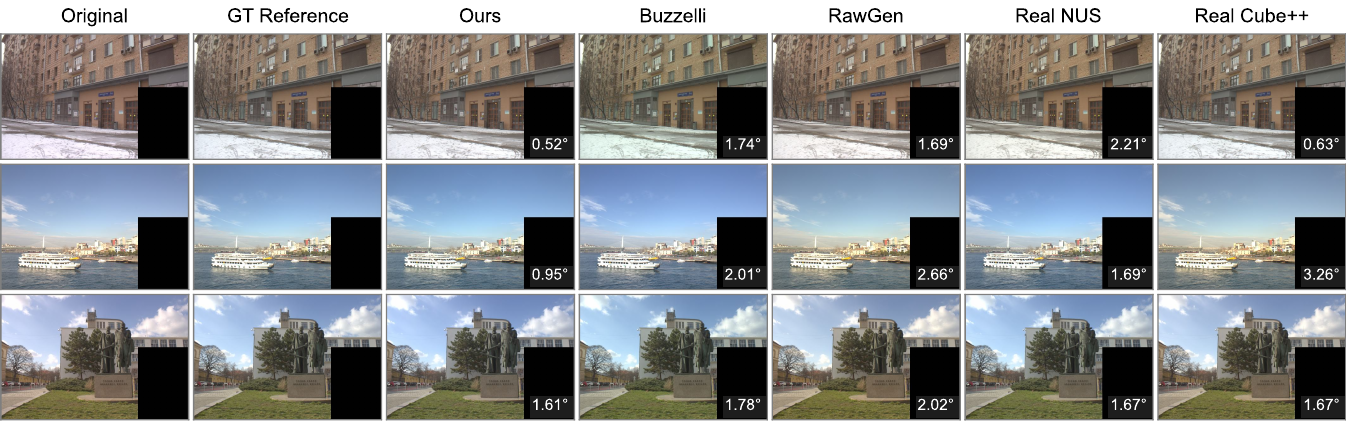}
    \caption{\textbf{Qualitative color constancy results on Cube++.}
    Three test images corrected using FC4 models trained
    on different data sources.
    Columns show the original image, ground-truth
    white-balance reference, and predictions from models
    trained on synthetic or real data.
    Values indicate per-image illuminant angular errors
    in degrees (lower is better).
    Real Cube++ is the in-domain reference.}
    \label{fig:awb_qualitative}
\end{figure*}

Color constancy models are trained on RAW images of one camera, so each new camera needs its own training data.
This experiment checks that RAW images projected from our HSIs can serve as such data, next to existing generators of synthetic RAW images.
We train color constancy models on RAW images from three synthetic sources, \ours, RawGen~\cite{kim2026rawgen} and Buzzelli \etal~\cite{buzzelli2023pipeline}, and use training on real images of the same camera as an in-domain reference.

For every rendered view, we also render an illumination map that records the spectral irradiance arriving at each visible surface point.
It is obtained by replacing, in \cref{eq:framework/spectral_rendering}, the material at the first camera-ray intersection with a white Lambertian reflector, while later bounces keep the scene materials, so the map includes light reflected by colored surfaces.
The methods we evaluate estimate one illuminant per image, so we project each pixel of the illumination map to RAW through the SSFs of the camera, as we do for the rendered images, and take the median over the map as the ground-truth illuminant.

For RawGen, whose original training data and prompts are not public, we use the released checkpoint to generate 3,000 RAW images from text prompts and build the training data as in RawGen~\cite{kim2026rawgen} and Graphics2RAW~\cite{seo2023graphics2raw}.
Each image is converted to the RAW space of the camera under an illuminant sampled from the released illuminant set, and this illuminant serves as the label.
For Buzzelli \etal, we follow their pipeline~\cite{buzzelli2023pipeline}, which converts each sRGB image to reflectance spectra and renders it under one illuminant spectrum.
The rendered image and the illuminant spectrum are projected to RAW through the same SSFs, and the projected illuminant is the ground truth.

We evaluate four color constancy methods, ConvMean~\cite{gong2019convmean}, FC4~\cite{hu2017fc4}, FFCC~\cite{barron2017ffcc} and SIIE~\cite{afifi2019siie}, on two real datasets, NUS~\cite{cheng2014nus} and Cube++~\cite{ershov2020cube}, using only images of the Canon EOS 600D, whose SSFs come from CamSpec~\cite{jiang2013camspec}.
For each method and training source, we train a separate model from scratch, select it on the validation set of its training source, and test it on real images without fine-tuning.
The in-domain reference is trained on the official training split of the dataset it is tested on, and all models are tested on the official test splits of NUS and Cube++.
We report the angular error between the estimated and the ground-truth illuminant in camera RGB space.
NUS gives one ground-truth illuminant per image, and Cube++ gives two, from the left and right sides of a SpyderCube target~\cite{ershov2020cube}.
Following Cogo \etal~\cite{cogo2026dense}, the error on Cube++ is the mean of two angular errors, one between the estimate and the left-side illuminant and one between the estimate and the right-side illuminant.

\begin{table}[!ht]
\centering
\caption{\textbf{Color constancy on real Canon EOS 600D images.}
Models are trained from scratch on each source (rows) and tested on the official test splits of NUS and Cube++ (columns).
Values are mean angular errors in degrees.
Real is an in-domain reference, trained and tested on the same dataset, and is not ranked.
Bold and underline mark the best and second-best synthetic source per column.}
\label{tab:awb}
\footnotesize
\renewcommand{\arraystretch}{1.6}
\setlength{\tabcolsep}{2.3pt}
\begin{tabular}{l cccc cccc}
\toprule
& \multicolumn{4}{c}{NUS} & \multicolumn{4}{c}{Cube++} \\
\cmidrule(lr){2-5}\cmidrule(lr){6-9}
Train set & ConvMean & FC4 & FFCC & SIIE & ConvMean & FC4 & FFCC & SIIE \\
\midrule
RawGen   & 3.83 & 3.80 & \underline{3.01} & 3.82 & \underline{4.23} & 3.34 & \underline{2.88} & \underline{2.79} \\
Buzzelli & \textbf{3.05} & \textbf{2.23} & \textbf{2.99} & \underline{3.23} & 5.32 & \underline{2.72} & 2.99 & 3.01 \\
Ours     & \underline{3.39} & \underline{2.75} & 3.18 & \textbf{2.94} & \textbf{3.39} & \textbf{2.33} & \textbf{2.60} & \textbf{2.56} \\
\midrule
Real & 2.52 & 2.39 & 2.64 & 2.14 & 2.05 & 1.78 & 2.43 & 1.92 \\
\bottomrule
\end{tabular}
\end{table}

\Cref{tab:awb} shows no clear winner among the synthetic sources on these two datasets, since the best source changes with the method and the dataset.
Our data ranks first or second in all cases but one.
\Cref{fig:awb_qualitative} shows examples on Cube++.

\section{Conclusion}

We presented \ours, a framework that converts existing 3D scene datasets into hyperspectral images by assigning measured reflectance spectra by material category and rendering each scene by spectral path tracing.
A hyperspectral reconstruction network trained only on our data generalizes to unseen real datasets better than networks trained on real HSIs on most test sets, and assigning spectra by material category instead of by color alone improves this transfer on every test set.
For color constancy, RAW images projected from our HSIs are useful training data, performing on par with existing synthetic RAW generators.
Because RAW images follow from the HSIs by projection, the same rendered data can serve any camera with known SSFs without new captures.
These SSFs, however, are costly to obtain~\cite{jiang2013camspec,seo2023graphics2raw}.
Combining our data with camera-agnostic methods, which transfer to new cameras without per-camera training data~\cite{afifi2019siie,afifi2021c5}, is a natural next step, as is multi-illuminant estimation from the rendered per-pixel illumination maps.
We release the 4,500 rendered HSIs, the code of \ours and its material-labeled reflectance library.

\clearpage
{ \small \bibliographystyle{ieeenat_fullname} \bibliography{main} }

\appendix
\section*{Technical Appendices and Supplementary Material}
\section{Material-Labeled Reflectance Library}
\label{supp:library}

\paragraph{Labeling hyperspectral images.}\label{supp:sec:labeling}
Most close-range hyperspectral datasets provide no material labels, so we label their pixels automatically with Segment Anything Model~3 (SAM~3)~\cite{framework/sam3}.
Because SAM~3 works on RGB images, each hyperspectral image is first rendered to sRGB from its 400--700\,nm range under the CIE D65 illuminant.
SAM~3 then segments this image once for every material category.
The text prompt is the category name, extended with additional terms for some categories.
Only masks with a confidence above 0.8 are kept.
Spectrophotometer measurements already carry material names, which we map to our categories separately for each source table.

\paragraph{Material categories.}\label{supp:sec:categories}
The library defines 63 material categories, listed here by their identifiers:

\smallskip{\raggedright\small
\texttt{asphalt}, \texttt{bark}, \texttt{bone}, \texttt{brick}, \texttt{cardboard}, \texttt{carpet}, \texttt{ceiling\_tile}, \texttt{ceramic}, \texttt{chalkboard}, \texttt{concrete}, \texttt{cork}, \texttt{dust}, \texttt{engineered\_stone}, \texttt{fabric}, \texttt{fabric\_lampshade}, \texttt{fire}, \texttt{foliage}, \texttt{food}, \texttt{fur}, \texttt{gemstone}, \texttt{glass}, \texttt{glass\_window}, \texttt{granite}, \texttt{grass}, \texttt{gravel}, \texttt{hair}, \texttt{ice}, \texttt{laminate}, \texttt{leather}, \texttt{linoleum}, \texttt{marble}, \texttt{metal}, \texttt{metal\_corroded}, \texttt{mirror}, \texttt{painted}, \texttt{paper}, \texttt{paper\_printed}, \texttt{plaster}, \texttt{plastic\_clear}, \texttt{plastic\_glossy}, \texttt{plastic\_opaque}, \texttt{polished\_stone}, \texttt{roofing\_tile}, \texttt{rubber}, \texttt{sand}, \texttt{skin}, \texttt{sky}, \texttt{snow}, \texttt{soil}, \texttt{sponge}, \texttt{stone}, \texttt{styrofoam}, \texttt{terracotta}, \texttt{tile}, \texttt{wallpaper}, \texttt{water}, \texttt{wax}, \texttt{whiteboard}, \texttt{wicker}, \texttt{wood}, \texttt{wood\_composite}, \texttt{wood\_floor}, \texttt{wood\_varnished}.\par}

\smallskip\noindent Where the category name alone would miss typical objects made of the material, the category receives additional prompts.
For example, \texttt{foliage} is also prompted with \texttt{leaves}, \texttt{tree}, \texttt{potted plant}, \texttt{bush} and \texttt{grass}.
\Cref{tab:supp-categories} gives the number of spectra, the number of source datasets and the color coverage of the categories.
Some categories are kept in the library but not used for rendering, for example \texttt{metal}, because metals are rendered as rough conductors with measured optical constants (\cref{sec:invscene}).

\begin{table*}[!ht]
  \centering
  \caption{\textbf{Material categories of the library.}
    Number of spectra, number of source datasets and chromaticity coverage of every material category that contains spectra.
    Cov.\ is the part of the $xy$ chromaticity diagram covered by the colors of the category (\cref{supp:library}).}
  \label{tab:supp-categories}
  \small
  \setlength{\tabcolsep}{4pt}
  \resizebox{\textwidth}{!}{%
  \begin{tabular}{@{}lrrr c lrrr c lrrr@{}}
    \toprule
    Category & Spectra & Datasets & Cov., \% && Category & Spectra & Datasets & Cov., \% && Category & Spectra & Datasets & Cov., \% \\
    \cmidrule(r){1-4} \cmidrule(lr){6-9} \cmidrule(l){11-14}
    \texttt{asphalt} & 15.9M & 11 & 18.2 && \texttt{granite} & 900.7K & 9 & 6.3 && \texttt{sand} & 696.0K & 7 & 6.2 \\
    \texttt{bark} & 1.3M & 10 & 12.8 && \texttt{grass} & 3.5M & 10 & 26.2 && \texttt{skin} & 2.6M & 15 & 24.6 \\
    \texttt{bone} & 327.8K & 7 & 8.7 && \texttt{gravel} & 325.0K & 5 & 6.4 && \texttt{sky} & 14.5M & 13 & 11.0 \\
    \texttt{brick} & 13.2M & 11 & 11.1 && \texttt{hair} & 463.9K & 14 & 21.5 && \texttt{snow} & 6.3M & 6 & 4.2 \\
    \texttt{cardboard} & 602.1K & 10 & 26.1 && \texttt{ice} & 56.1K & 2 & 1.1 && \texttt{soil} & 3.2M & 11 & 16.5 \\
    \texttt{carpet} & 730.5K & 6 & 25.6 && \texttt{laminate} & 30.0K & 1 & 1.3 && \texttt{sponge} & 34.6K & 3 & 5.0 \\
    \texttt{ceiling\_tile} & 30.0K & 2 & 0.5 && \texttt{leather} & 148.9K & 5 & 7.3 && \texttt{stone} & 11.0M & 22 & 20.8 \\
    \texttt{ceramic} & 17.2M & 8 & 20.2 && \texttt{marble} & 488.0K & 5 & 5.9 && \texttt{styrofoam} & 57.8K & 4 & 1.7 \\
    \texttt{chalkboard} & 25.1K & 3 & 2.6 && \texttt{metal} & 7.7M & 20 & 24.2 && \texttt{terracotta} & 1.0M & 11 & 23.0 \\
    \texttt{concrete} & 10.0M & 12 & 10.6 && \texttt{metal\_corroded} & 1.6M & 10 & 18.6 && \texttt{tile} & 5.9M & 13 & 23.6 \\
    \texttt{cork} & 50.0K & 3 & 2.9 && \texttt{mirror} & 554.1K & 7 & 11.0 && \texttt{wallpaper} & 72.5K & 1 & 8.8 \\
    \texttt{dust} & 25 & 1 & 0.1 && \texttt{painted} & 2.4M & 21 & 26.9 && \texttt{water} & 1.0M & 12 & 17.6 \\
    \texttt{engineered\_stone} & 108.8K & 2 & 1.8 && \texttt{paper} & 2.7M & 15 & 25.9 && \texttt{wax} & 292.3K & 5 & 16.0 \\
    \texttt{fabric} & 7.6M & 21 & 43.8 && \texttt{paper\_printed} & 563.4K & 9 & 20.4 && \texttt{whiteboard} & 124.0K & 4 & 2.7 \\
    \texttt{fabric\_lampshade} & 55.8K & 6 & 4.7 && \texttt{plaster} & 351.2K & 4 & 3.5 && \texttt{wicker} & 15.1K & 2 & 4.8 \\
    \texttt{foliage} & 18.5M & 36 & 52.4 && \texttt{plastic\_clear} & 3.1M & 11 & 19.4 && \texttt{wood} & 3.3M & 18 & 31.2 \\
    \texttt{food} & 183.0M & 29 & 35.1 && \texttt{plastic\_glossy} & 265.1K & 5 & 27.6 && \texttt{wood\_composite} & 4.8M & 27 & 26.2 \\
    \texttt{gemstone} & 127.6K & 4 & 14.5 && \texttt{plastic\_opaque} & 7.9M & 16 & 43.2 && \texttt{wood\_floor} & 603.2K & 8 & 10.3 \\
    \texttt{glass} & 5.8M & 11 & 31.2 && \texttt{polished\_stone} & 519.9K & 4 & 4.2 && \texttt{wood\_varnished} & 401.0K & 8 & 10.4 \\
    \texttt{glass\_window} & 6.4M & 10 & 23.3 && \texttt{rubber} & 2.0M & 15 & 38.2 &&  & & &  \\
    \bottomrule
  \end{tabular}}
\end{table*}

\paragraph{Color coverage.}\label{supp:sec:coverage}
We show the colors that a category spans by projecting every spectrum of the category to CIE~1931 $xy$ chromaticity under D65.
The coverage of a category is the share of the area inside the spectral locus that its chromaticities reach.
For reference, the sRGB gamut covers 33.7\% of the locus, and all categories together cover 60.8\%.
A single category is not expected to cover the whole locus, since many materials occur only in a limited range of colors.
Coverage also depends on the spectra available for each category, so it describes what the library contains rather than the full color range of a material.

\begin{figure*}[t]
\centering
\captionsetup[sub]{font=footnotesize,labelformat=empty}
\newcommand{\locus}[2]{\begin{subfigure}[t]{0.195\linewidth}\centering\includegraphics[width=\linewidth]{fig/supp/locus_#1}\caption{#2}\end{subfigure}}
\locus{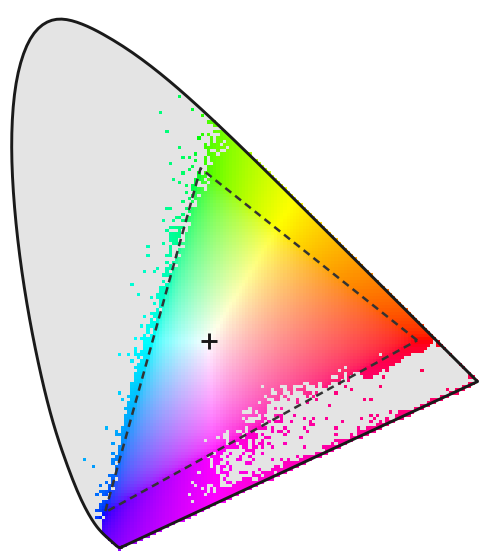}{\texttt{foliage} (52.4\%)}\hspace{0.015\linewidth}
\locus{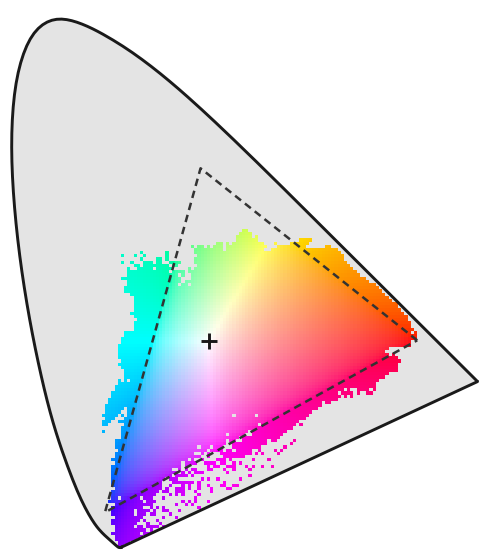}{\texttt{fabric} (43.8\%)}\hspace{0.015\linewidth}
\locus{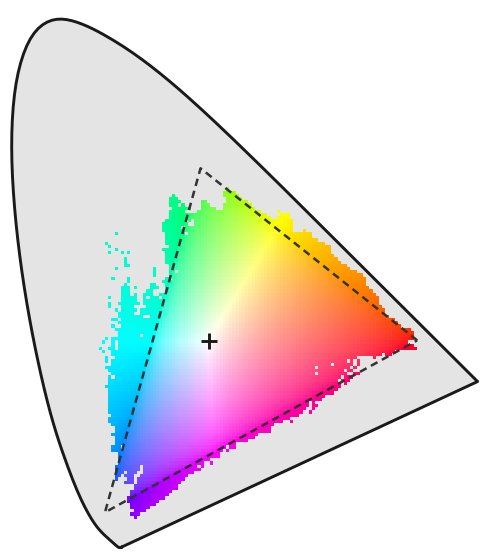}{\texttt{plastic\_opaque} (43.2\%)}\hspace{0.015\linewidth}
\locus{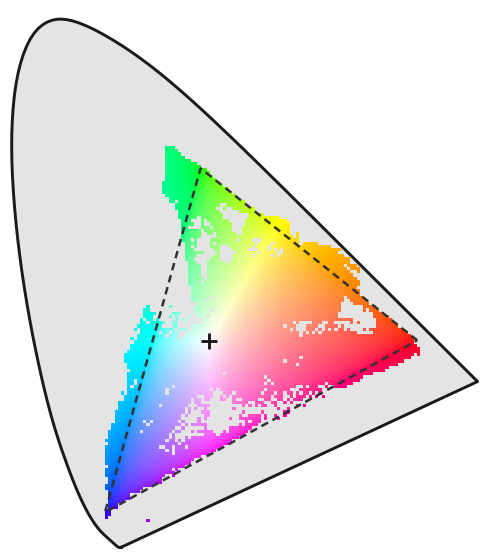}{\texttt{rubber} (38.2\%)}\\[2pt]
\locus{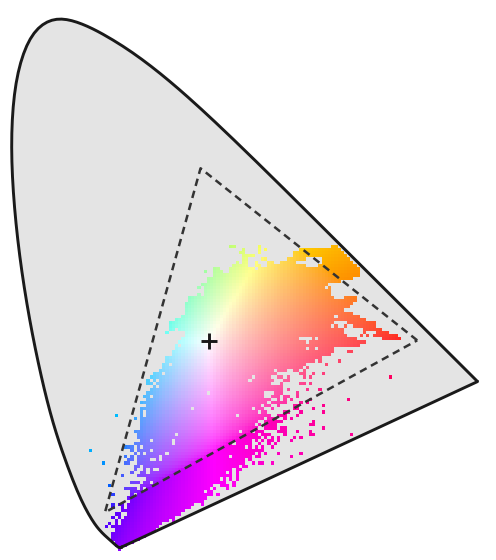}{\texttt{wood} (31.2\%)}\hspace{0.015\linewidth}
\locus{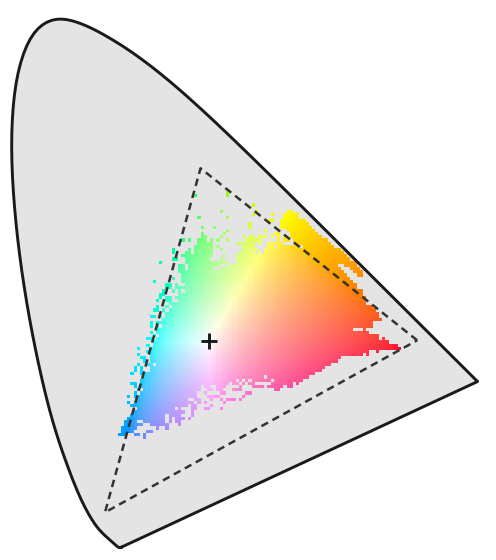}{\texttt{painted} (26.9\%)}\hspace{0.015\linewidth}
\locus{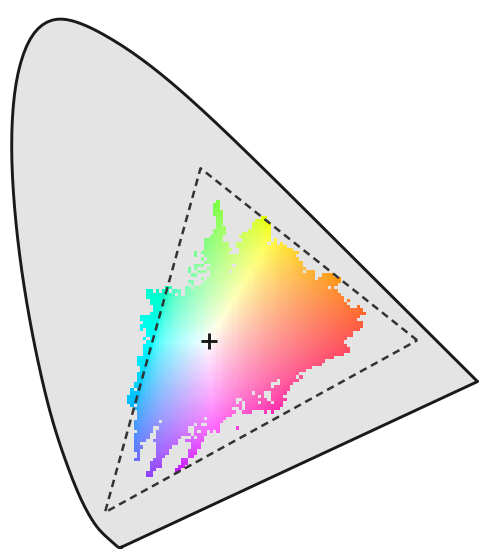}{\texttt{paper} (25.9\%)}\hspace{0.015\linewidth}
\locus{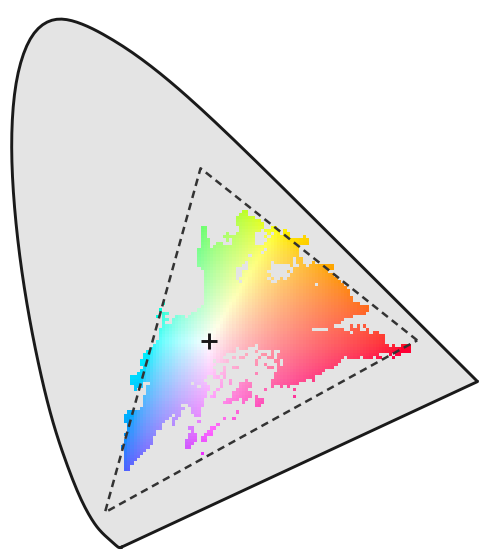}{\texttt{tile} (23.6\%)}
\caption{\textbf{Chromaticity coverage of material categories.}
Each panel shows the $xy$ chromaticities reached by one category, with its coverage in brackets.
The dashed triangle marks the sRGB gamut.}
\label{fig:supp-locus}
\end{figure*}

\Cref{fig:supp-locus} shows the coverage of eight categories.
The categories \texttt{foliage}, \texttt{fabric}, \texttt{plastic\_opaque} and \texttt{rubber} reach 38--52\% of the locus, more than the sRGB gamut, while \texttt{wood}, \texttt{painted}, \texttt{paper} and \texttt{tile} reach 24--31\%.
\Cref{tab:supp-categories} gives the coverage of every category.

\paragraph{Data availability.}
The library and the list of its source datasets will be released upon acceptance.

\section{Cross-Dataset Evaluation with Other HSR Methods}
\label{supp:hsr}

\providecommand{\TODO}[1]{\textcolor{red}{[TODO: #1]}}
\providecommand{\indom}{\cellcolor{gray!15}} %
\providecommand{\best}[1]{\textbf{#1}}
\providecommand{\second}[1]{\underline{#1}}

To check that the benefit of our data does not depend on the reconstruction method, we repeat the cross-dataset evaluation of \cref{sec:exp-hsi} with three HSR methods of other types, the dictionary-based A+~\cite{wu2017defense}, the CNN AWAN~\cite{li2020adaptive} and the Mamba-based GMSR~\cite{wang2025gmsr}, in addition to the transformer-based MST++~\cite{cai2022mstpp} of \cref{tab:cross_dataset}.
Each method keeps the model hyperparameters of its original paper, and the data, RAW rendering, splits and metrics are those of \cref{sec:exp-hsi}.

\paragraph{RAW simulation.}
All HSIs are projected to RAW RGB through the spectral sensitivities of a Nikon D700~\cite{jiang2013camspec}, scaled so that the largest channel sum equals one.
We then add Poisson shot noise as in the NTIRE 2022 spectral recovery challenge~\cite{arad2022ntire}, with a full-well capacity of 750 photoelectrons.
The noisy signal is clipped at the saturation level and quantized to 12 bits.
Every training source and every test set goes through the same simulation.

\paragraph{Data splits.}\label{supp:sec:hsr-details}
Every dataset is split into training, validation and test partitions, and all images of one scene stay in the same partition.
The test labels of the ARAD~1K challenge~\cite{arad2022ntire} are not public, so we train on its 899 public training images and divide its 50 official validation images equally between our validation and test partitions.
ICVL~\cite{arad2016sparse} images are grouped by location and capture date, and whole groups are assigned to partitions so that validation and test each hold about 10\% of the images.
BJTU-UVA~\cite{du2025automatic} keeps its official split.
The data of Buzzelli \etal~\cite{buzzelli2023pipeline} keeps all renderings of one INTEL-TAU~\cite{laakom2021intel} scene, taken by different cameras, in the same partition, and our data is split by 3D scene, so all views of one scene fall into the same partition.

\paragraph{Results.}\label{supp:sec:hsr-models}
\Cref{tab:supp_xds} shows that the findings of \cref{sec:exp-hsi} hold for all four methods.
A model trained on one real dataset is worse on every other real dataset in all three metrics.
Training on our data gives the best or second-best out-of-domain result in every column for MST++, AWAN and GMSR, and it outperforms the synthetic HSIs of Buzzelli \etal in every cell for every method.
The advantage of our data therefore does not depend on the reconstruction method.

\providecommand{\indom}{\cellcolor{gray!15}}
\providecommand{\best}[1]{\textbf{#1}}
\providecommand{\second}[1]{\underline{#1}}
\begin{table*}[!ht]
  \centering
  \caption{\textbf{Cross-dataset generalization across HSR methods.}
    The table follows~\cref{tab:cross_dataset}, with one block per method (vertical label), and the MST++~\cite{cai2022mstpp} block repeats~\cref{tab:cross_dataset}.
    Bold marks the best and underline the second-best out-of-domain result per column within each method.
    Arrows show whether higher or lower is better.}
  \label{tab:supp_xds}
  \setlength{\tabcolsep}{3pt}
  \renewcommand{\arraystretch}{1.6}
  \resizebox{\textwidth}{!}{%
  \begin{tabular}{c l ccc ccc ccc ccc}
    \toprule
    & \multirow{2}{*}{Train set}
      & \multicolumn{3}{c}{ARAD~1K}
      & \multicolumn{3}{c}{Doomer}
      & \multicolumn{3}{c}{ICVL}
      & \multicolumn{3}{c}{BJTU-UVA} \\
    \cmidrule(lr){3-5} \cmidrule(lr){6-8} \cmidrule(lr){9-11} \cmidrule(lr){12-14}
      & & PSNR, dB$\uparrow$ & SAM, $^\circ$$\downarrow$ & NSE, \%$\downarrow$
        & PSNR, dB$\uparrow$ & SAM, $^\circ$$\downarrow$ & NSE, \%$\downarrow$
        & PSNR, dB$\uparrow$ & SAM, $^\circ$$\downarrow$ & NSE, \%$\downarrow$
        & PSNR, dB$\uparrow$ & SAM, $^\circ$$\downarrow$ & NSE, \%$\downarrow$ \\
    \midrule
    \multirow{6}{*}{\rotatebox[origin=c]{90}{MST++~\cite{cai2022mstpp}}} & ARAD~1K
      & \indom 40.44 & \indom 3.90 & \indom 7.87
      & 27.88 & 13.42 & 25.31
      & \second{37.47} & \best{3.74} & \best{5.57}
      & 30.53 & 10.08 & 15.68 \\
    & Doomer
      & \second{33.61} & \second{8.71} & \second{13.11}
      & \indom 35.90 & \indom 4.54 & \indom 8.21
      & 36.29 & 4.65 & 6.69
      & 31.42 & \second{9.40} & \second{15.01} \\
    & ICVL
      & 32.99 & 11.32 & 23.46
      & 27.14 & 14.03 & 30.15
      & \indom 43.83 & \indom 1.71 & \indom 3.10
      & \best{32.86} & 9.45 & 19.02 \\
    & BJTU-UVA
      & 30.98 & 11.50 & 15.29
      & 26.50 & 13.84 & 22.93
      & 33.33 & 7.67 & 10.13
      & \indom 42.67 & \indom 2.11 & \indom 4.72 \\
    \cmidrule(lr){2-14}
    & Buzzelli
      & 29.98 & 12.79 & 16.85
      & \second{28.08} & \second{11.29} & \second{18.27}
      & 33.97 & 6.50 & 9.47
      & 27.16 & 16.21 & 25.25 \\
    & Ours
      & \best{36.99} & \best{6.54} & \best{10.54}
      & \best{28.19} & \best{11.21} & \best{18.19}
      & \best{37.55} & \second{4.13} & \second{6.12}
      & \second{31.51} & \best{8.79} & \best{13.69} \\
    \specialrule{0.7pt}{4pt}{4pt}
    \multirow{6}{*}{\rotatebox[origin=c]{90}{AWAN~\cite{li2020adaptive}}} & ARAD~1K
      & \indom 40.13 & \indom 4.16 & \indom 7.34
      & \second{28.21} & \second{12.67} & \second{20.60}
      & \second{37.01} & \second{4.07} & \best{5.77}
      & 29.75 & 10.52 & 15.81 \\
    & Doomer
      & 30.77 & 11.57 & 16.27
      & \indom 36.01 & \indom 4.13 & \indom 6.73
      & 31.81 & 7.27 & 10.13
      & 31.14 & \second{9.47} & \second{14.92} \\
    & ICVL
      & \second{32.84} & 11.58 & 17.83
      & 26.79 & 14.39 & 23.29
      & \indom 44.42 & \indom 1.47 & \indom 2.58
      & \second{32.08} & 10.14 & 15.47 \\
    & BJTU-UVA
      & 31.22 & \second{11.30} & \best{14.88}
      & 26.04 & 13.94 & 22.18
      & 32.59 & 8.16 & 10.75
      & \indom 43.21 & \indom 2.03 & \indom 4.38 \\
    \cmidrule(lr){2-14}
    & Buzzelli
      & 26.84 & 21.13 & 26.76
      & 25.96 & 15.34 & 22.29
      & 25.44 & 19.20 & 24.52
      & 25.33 & 18.95 & 29.07 \\
    & Ours
      & \best{33.32} & \best{11.09} & \second{15.78}
      & \best{28.51} & \best{11.12} & \best{19.22}
      & \best{38.10} & \best{4.00} & \second{5.90}
      & \best{32.59} & \best{7.41} & \best{11.77} \\
    \specialrule{0.7pt}{4pt}{4pt}
    \multirow{6}{*}{\rotatebox[origin=c]{90}{GMSR~\cite{wang2025gmsr}}} & ARAD~1K
      & \indom 38.04 & \indom 5.42 & \indom 9.91
      & \best{27.94} & \second{13.47} & 26.10
      & \second{36.73} & \second{4.25} & \best{6.64}
      & 32.10 & \second{8.95} & \second{12.83} \\
    & Doomer
      & 31.98 & 11.71 & 17.60
      & \indom 32.35 & \indom 7.53 & \indom 13.17
      & 33.85 & 5.85 & 8.93
      & 31.08 & 9.42 & 16.21 \\
    & ICVL
      & \second{33.02} & \second{9.93} & 21.10
      & 26.59 & 15.32 & 32.62
      & \indom 40.54 & \indom 2.60 & \indom 4.58
      & \second{32.40} & 9.74 & 17.87 \\
    & BJTU-UVA
      & 30.68 & 11.95 & \second{16.52}
      & 25.91 & 14.76 & 23.84
      & 32.29 & 8.73 & 11.77
      & \indom 40.71 & \indom 2.84 & \indom 5.97 \\
    \cmidrule(lr){2-14}
    & Buzzelli
      & 28.38 & 16.91 & 21.49
      & 26.38 & 14.13 & \second{21.78}
      & 29.78 & 10.17 & 13.61
      & 26.39 & 16.88 & 26.24 \\
    & Ours
      & \best{33.67} & \best{8.84} & \best{13.76}
      & \second{27.88} & \best{11.55} & \best{19.09}
      & \best{36.78} & \best{3.88} & \second{6.83}
      & \best{33.17} & \best{7.21} & \best{11.74} \\
    \specialrule{0.7pt}{4pt}{4pt}
    \multirow{6}{*}{\rotatebox[origin=c]{90}{A+~\cite{wu2017defense}}} & ARAD~1K
      & \indom 32.80 & \indom 8.40 & \indom 16.95
      & \second{25.22} & \second{15.29} & \second{27.36}
      & \best{32.35} & \second{7.09} & \second{12.39}
      & 26.42 & 14.57 & 24.62 \\
    & Doomer
      & \best{31.67} & \second{11.92} & \best{19.87}
      & \indom 28.39 & \indom 11.45 & \indom 20.58
      & \second{32.35} & 7.51 & 12.49
      & \best{29.48} & \best{11.60} & \best{20.63} \\
    & ICVL
      & 30.19 & 15.09 & 26.20
      & 25.19 & 15.95 & 29.05
      & \indom 34.03 & \indom 4.81 & \indom 9.59
      & 26.74 & 14.65 & 26.68 \\
    & BJTU-UVA
      & 27.73 & 16.33 & 25.26
      & 22.92 & 18.98 & 31.40
      & 28.26 & 12.88 & 19.18
      & \indom 34.32 & \indom 5.60 & \indom 11.58 \\
    \cmidrule(lr){2-14}
    & Buzzelli
      & 26.94 & 20.65 & 29.52
      & 24.10 & 17.55 & 27.78
      & 26.60 & 14.89 & 21.97
      & 28.66 & 13.10 & 21.11 \\
    & Ours
      & \second{31.22} & \best{11.64} & \second{20.23}
      & \best{26.74} & \best{13.23} & \best{23.33}
      & 31.84 & \best{6.97} & \best{12.26}
      & \second{28.88} & \second{11.98} & \second{20.88} \\
    \bottomrule
  \end{tabular}%
  }
\end{table*}

\end{document}